\documentclass[journal]{IEEEtran}
\usepackage{graphicx}
\usepackage{subfigure} 
\usepackage{hyperref}

\usepackage{graphicx}

\usepackage{caption}
\usepackage{multirow}
\usepackage{amsmath}
\usepackage{url}
\usepackage{xcolor}
\usepackage{hyperref}
\usepackage{booktabs}
\usepackage{float}
\usepackage{amssymb}
\usepackage{algorithm}
\usepackage[numbers,sort&compress]{natbib}

\ifCLASSINFOpdf
\else
\fi
\begin{document}
%
\title{ACTIVE: A Deep Model for Sperm and Impurity Detection in Microscopic Videos}
%
%
%

\author{Ao~Chen,
        Jinghua~Zhang,
        Md Mamunur Rahaman,
        Hongzan~Sun, \textit{M.D.},
        Tieyong Zeng,
        Marcin~Grzegorzek,
        Feng-Lei~Fan$^*$, \textit{IEEE Member},
        Chen~Li$^*$
\thanks{$^*$Feng-Lei Fan (fengleifan@cuhk.edu.hk) and Chen Li (lichen@bmie.neu.edu.cn) are co-corresponding authors. 
This work is supported by NSFC Fund No. 82220108007.}
\thanks{Ao~Chen, Md Mamunur Rahaman, and Chen~Li are with the Microscopic Image and Medical Image Analysis Group, 
College of Medicine and Biological Information Engineering, Northeastern University, Shenyang, PR China.}
\thanks{Jinghua~Zhang is with College of Intelligence Science and Technology, National University of Defense Technology, PR China.}
\thanks{Hongzan~Sun (M.D.) is with Shengjing Hospital, China Medical University, Shenyang, PR China.}
\thanks{Marcin Grzegorzek is with Institute of Medical Informatics, University of Luebeck, Germany}
\thanks{Tieyong Zeng and Feng-Lei~Fan is with Department of Mathematics, The Chinese University of Hong Kong, Shatin, N.T. Hong Kong.}
}

\markboth{Journal of \LaTeX\ Class Files,~Vol.~**, No.~**, May~2023}%
{Shell \MakeLowercase{\textit{et al.}}: Bare Demo of IEEEtran.cls for IEEE Journals\cite{iammarrone2003male}}

\maketitle

\begin{abstract}
The accurate detection of sperms and impurities is a very challenging task, facing problems such as the small 
size of targets, indefinite target morphologies, low contrast and resolution of the video, and similarity of 
sperms and impurities. So far, the detection of sperms and impurities still largely relies on the traditional 
image processing and detection techniques which only yield limited performance and often require manual 
intervention in the detection process, therefore unfavorably escalating the time cost and injecting the subjective 
bias into the analysis. Encouraged by the successes of deep learning methods in numerous object detection tasks, 
here we report a deep learning model based on \emph{Double Branch Feature Extraction Network} (DBFEN) and 
\emph{Cross-conjugate Feature Pyramid Networks} (CCFPN). DBFEN is designed to extract visual features from tiny 
objects with a double branch structure, and CCFPN is further introduced to fuse the features extracted by DBFEN 
to enhance the description of position and high-level semantic information. Our work is the pioneer of introducing 
deep learning approaches to the detection of sperms and impurities. Experiments show that the highest 
${\rm AP}_{50}$ of the sperm and impurity detection is 91.13\% and 59.64\%, which lead its competitors by a 
substantial margin and establish new state-of-the-art results in this problem.

\end{abstract}

\begin{IEEEkeywords}
Semen analysis, sperm microscopy videos, object detection, deep learning.
\end{IEEEkeywords}

\IEEEpeerreviewmaketitle

\section{Introduction}

\begin{figure}[!htbp]
\centering
\includegraphics[width=0.95\linewidth]{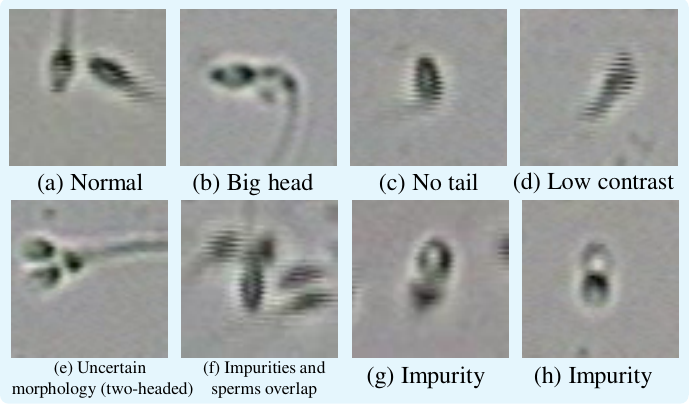}
\caption{Examples of different sperm types.}
\vspace{-0.2cm}
\label{FIG:1}
\end{figure}

\IEEEPARstart{I}{nfertility} affects 15\% of couples worldwide~\cite{maharlouei2021prevalence}. Based on clinical 
studies, the male factor takes up around 20\% of all infertility cases~\cite{fallara2021male}, and most of them 
are related to the quality of semen~\cite{kumar2015trends}. Currently, semen analysis is one of the most essential 
and effective techniques to investigate the male infertility. \emph{Computer-Aided Semen Analysis} (CASA) based 
on sperm microscopy videos is widely used in semen analysis because of its ability to provide an objective evaluation 
of a large number of sperms in a short time~\cite{zhao2022survey}. The first and foremost step in the CASA system 
is the detection of sperms and impurities, whose accuracy has a major impact on the downstream sperm evaluation. 
However, the accurate detection of sperms and impurities is a very challenging task, subjected to many difficulties 
such as the small size of targets, indefinite target morphologies, low contrast and resolution of the video, and 
similarity of sperms and impurities, as shown in Fig.~\ref{FIG:1}.

So far, the detection of sperms and impurities still largely relies on traditional image processing 
techniques~\cite{elsayed2015development,urbano2016automatic,zhou2009efficient,li2020foldover,yang2014head} which 
primarily include threshold-based methods, shape fitting methods, and filtering methods. 
\textbf{i)} Threshold-based methods find the best threshold to binarize an image~\cite{bhargavi2014survey}. 
In~\cite{elsayed2015development}, sperm detection is accomplished by sequentially performing contrast enhancement, 
grayscale conversion, background identification, and background subtraction operations on the image, and finally 
binarizing the image using maximum entropy as the best threshold. In~\cite{urbano2016automatic}, sperm detection 
is performed by sequentially performing Gaussian filtering and Laplacian-of-Gaussian (LoG) filtering operations 
on the image, and finally binarizing the image using the Ostu~\cite{otsu1979threshold} threshold method. 
\textbf{ii)} The shape-fitting methods use a polygon to fit the object. In~\cite{zhou2009efficient} and~\cite{yang2014head}, 
a rectangular area and an ellipse similar to the shape of the sperm are used to fit sperms, respectively, and then the 
parameters of the rectangle and ellipse are used to describe the position of the sperm, respectively. 
\textbf{iii)} The filtering methods filter the image with a suitable filter to separate the object and the background. 
In~\cite{ravanfar2011low}, morphological filters are used to isolate sperm cells. First, sperm cells are separated 
from other debris by filtering the image sequence according to a top-hat operation using appropriate structuring 
elements, followed by an open filter to reduce the remaining noisy objects smaller than the sperm head. However, 
these traditional detection methods only yield limited performance and often require manual intervention. 
Consequently, it not only escalates the time cost but also unavoidably injects the subjective bias into a CASA system. 
\begin{figure*}[!htbp]
	\centering
		\includegraphics[width=0.98\linewidth]{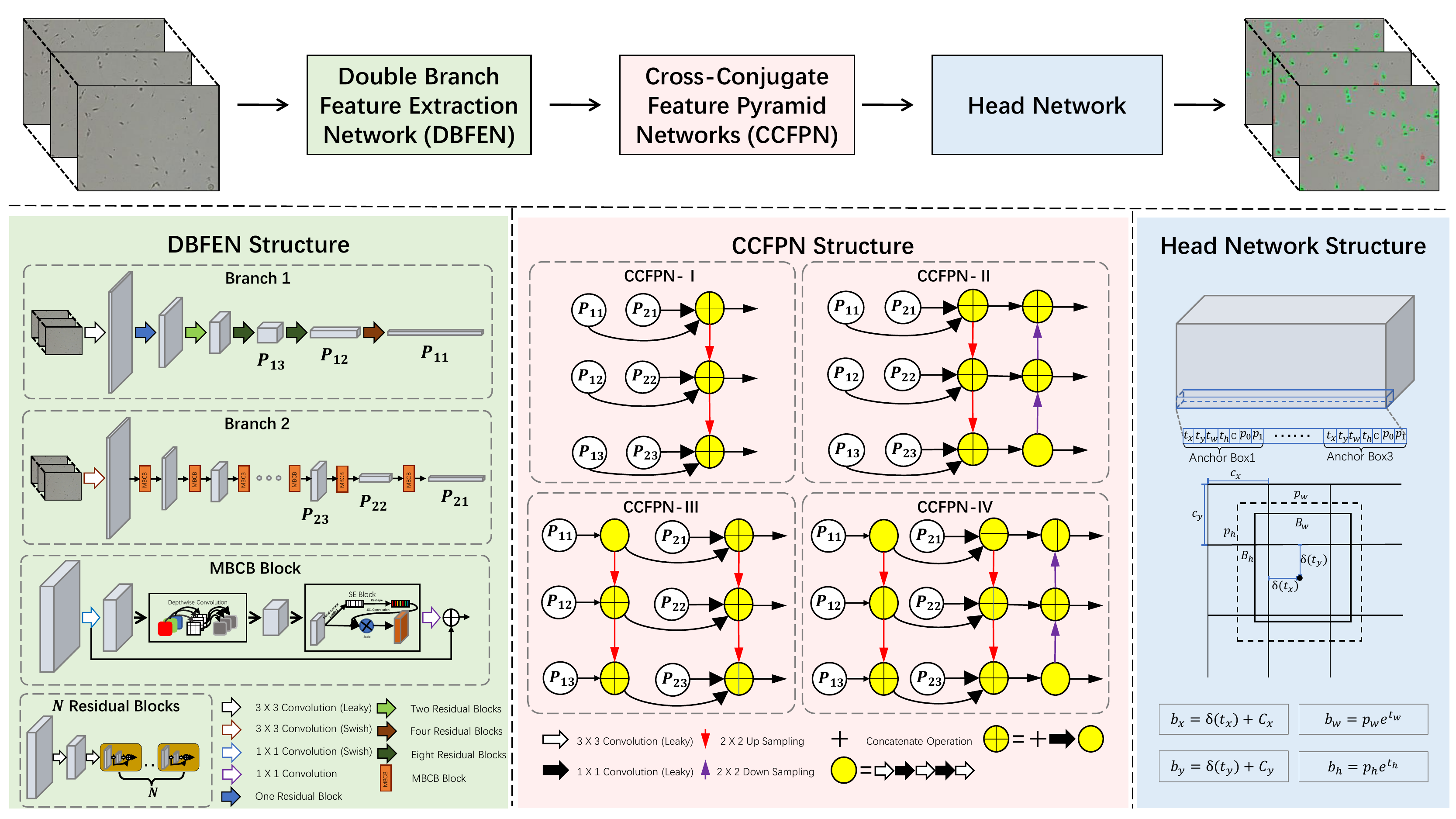}
	\caption{The architecture of ACTIVE. From the left to the right are the DBFEN, CCFPN and head network structures.}
 \vspace{-0.5cm}
	\label{FIG:2}
\end{figure*}

In recent years, deep learning (DL), a representative data-driven approach, has been dominating numerous 
imaging applications~\cite{wang2022ctformer}, ranging from low-level to high-level tasks. It has been widely 
confirmed by practitioners that provided well-curated big data and sufficient computing power, deep learning 
can deliver satisfactory results due to its strong feature extraction power. With the advent of the deep 
learning era, many excellent object detection models have been proposed, such as 
\emph{Region-based Convolutional Neural Networks} (RCNNs)~\cite{girshick2015fast,He2017mask}, 
\emph{You Only Look Once} (YOLO) series~\cite{redmon2018yolov3,bochkovskiy2020yolov4}, 
\emph{Single Shot Multibox Detector} (SSD)~\cite{liu2016ssd}, and Efficientdet~\cite{tan2020efficientdet}. 
It has been shown that CNNs can outperform classical image processing algorithms in most object detection 
tasks~\cite{zou2019object}. However, deep learning-based methods were unfortunately little explored 
in the field of sperm detection and analysis~\cite{zhao2022survey}. In~\cite{nissen2017convolutional}, 
a CNN network was designed to detect sperm, but there is a drawback in choosing the best threshold in the 
whole process. \cite{somasundaram2021faster} was mainly concerned with the sperm classification task. 
In~\cite{ZOU2022105543}, a deep learning method called TOD-CNN was proposed which obtains good detection 
performance for sperms but fails impurities.

In this investigation, we propose a deep learning model, referred to as ACTIVE (\underline{A} Deep Model for 
Sperm Dete\underline{CTI}on in Microscopic \underline{V}id\underline{E}os). As Fig.~\ref{FIG:2} shows, 
given the sperm videos from the \emph{Sperm Videos and Images Analysis} (SVIA) dataset~\cite{chen2022svia}, 
the proposed ACTIVE framework consists of three main parts: 
\emph{Double Branch Feature Extraction Network} (DBFEN), 
\emph{Cross-Conjugate Feature Pyramid Networks} (CCFPN), 
and predicted network. Two branches of DBFEN respectively use residual structures and mobile inverted bottleneck 
convolution block to extract features and enhance the generalization ability, while DBFEN can fuse abundant 
features of different resolutions to facilitate the detection of tiny sperms and impurities. The contributions 
of this paper are as follows:
\begin{itemize}
\item We propose a double-branch deep learning framework, namely ACTIVE, which facilitates feature extraction 
and utilizes a series of novel structures to perform the feature fusion. Our work is the pioneer of introducing 
deep learning approaches into the field of computer-aided semen analysis.

\item Experiments demonstrate that our proposed ACTIVE can overcome the difficulties of the sperm and impurity 
detection such as tiny size and indefinite morphology to achieve the state-of-the-art performance.
\end{itemize}

\section{Methodology}

\subsection{Double Branch Feature Extraction Network (DBFEN)}
DBFEN uses two branches to extract effective features of sperm microscopic images, which pays more attention to 
the features with the largest amount of information. Meanwhile, it also supports increasing the depth of the 
network to extract more features to further improve the detection performance.

\underline{Branch 1} mainly uses residual structures to extract features and facilitate training. The residual 
structure consists of a $3\times3$ convolutional filter with a step size of 2, a $1\times1$ convolutional filter, 
and a $3\times3$ convolutional filter. The $3\times3$ convolutional filter with the step size of 2 can help the 
network to compress the width and height of feature maps, which is similar to the function of the pooling operation. 
The feature map generated by the $3\times3$ convolutional filter with the step size of 2 is concatenated with the 
feature map generated by the $3\times3$ convolutional filter. This concatenation can avoid the gradient vanishment 
problem, thereby allowing a deep structure.

\underline{Branch 2} mainly uses \emph{Mobile inverted Bottleneck Convolution Block} (MBCB)~\cite{tan2020efficientdet} 
to extract features. It first employs a 1 $\times$ 1 convolution filter on the input and changes the output 
channel dimension by the expansion ratio (if the expansion ratio is $k$, the channel dimension will grow by $k$ times. 
If the expansion ratio is 1, the 1 $\times$ 1 convolution filter and subsequent batch normalization and activation 
functions will be directly omitted). Second, the 3 $\times$ 3 depth-wise convolution filter is applied. Then, 
the \emph{Squeeze and Excitation} (SE) block is introduced to this branch. After that, to recover the original 
channel dimension, a 1 $\times$ 1 convolution filter is used. Finally, the drop operation and skip connection are 
introduced to copy the input feature map to the end of the structure~\cite{huang2016deep}, which can discard the 
weights between hidden layers according to a fixed probability instead of simply discarding hidden nodes and sample 
the weights of each node with a Gaussian distribution in the testing stage to effectively solve problems caused by 
model quantization~\cite{mobiny2021dropconnect}.

When we use the repeated MBCB, this block will perform connection deactivation and input jump connection. 
Connection deactivation is an operation similar to random deactivation, and the skip connection is added to 
combine the feature maps at the earlier and later parts of the model. Therefore, we can use different numbers 
of MBCB to eliminate the dependence between neural units and enhance the generalization 
ability~\cite{tan2020efficientdet}. Simultaneously, the SE Block~\cite{hu2018squeeze} utilizes the attention 
mechanism or gate control on the channel dimension, which allows the model to pay more attention to the channel 
features with the largest amount of information and suppress those unimportant channel features.

\vspace{-0.3cm}
\subsection{Cross-Conjugate Feature Pyramid Networks (CCFPN)}
To handle the detection of tiny sperms and impurities, abundant features of different resolutions obtained from 
the DBFEN are aggregated. Traditional top-down \emph{Feature Pyramid Networks} (FPN)~\cite{lin2017feature} is 
inherently limited by unidirectional information flow, while \emph{Pyramid Attention Networks} 
(PANet)~\cite{liu2018path} adds an additional bottom-up path. Based on FPN and PANet, as shown in Fig.~\ref{FIG:2}, 
we propose CCFPN to fuse multi-resolution features: $\mathbf{P}^{\rm  out}= \textit{f} ( \mathbf{P}^{\rm  in} )$, 
where $\mathbf{P}^{\rm  in}=( P_{11}^{\rm in}, ..., P_{ij}^{\rm in}, ... )$ is a list of multi-scale features, 
and $P_{ij}^{\rm in}$ represents the input feature of the preceding layer $j$ of the $i$-th branch. CCFPN has 
four variants:

\vspace{-0.2cm}
{\begin{equation}
\label{EQ1}
\left\{\resizebox{0.85\hsize}{!}{$
\begin{aligned}
     &{\rm CCFPN_{\uppercase\expandafter{\romannumeral1}}} (P_{1}^{\rm out}) = {\rm Conv} ( P_{11}^{\rm in} + P_{21}^{\rm in} ) &                      \\
     &{\rm CCFPN_{\uppercase\expandafter{\romannumeral1}}} (P_{2}^{\rm out}) = {\rm Conv} ( P_{12}^{\rm in} + P_{22}^{\rm in} + {\rm Up} ( {\rm CCFPN_{\uppercase\expandafter{\romannumeral1}}} (P_{1}^{out}) ) )&\\
     &{\rm CCFPN_{\uppercase\expandafter{\romannumeral1}}} (P_{3}^{\rm out}) = {\rm Conv} ( P_{13}^{\rm in} + P_{23}^{\rm in} + {\rm Up} ( {\rm CCFPN_{\uppercase\expandafter{\romannumeral1}}} (P_{2}^{\rm out}) ) )&
\end{aligned}$}
\right.
\end{equation}

\vspace{-0.2cm}
\begin{equation}
\label{EQ2}
\left\{\resizebox{0.85\hsize}{!}{$
\begin{aligned}
     & {\rm CCFPN_{\uppercase\expandafter{\romannumeral2}}} (P_{1}^{\rm out}) = {\rm Conv} (  {\rm CCFPN_{\uppercase\expandafter{\romannumeral1}}} (P_{1}^{\rm out}) +  {\rm Down} ( {\rm CCFPN_{\uppercase\expandafter{\romannumeral2}}} (P_{2}^{\rm out}) )  ) &                      \\
     &{\rm CCFPN_{\uppercase\expandafter{\romannumeral2}}} (P_{2}^{\rm out}) = {\rm Conv} ({\rm CCFPN_{\uppercase\expandafter{\romannumeral1}}} (P_{2}^{\rm out})+ {\rm Down} ({\rm CCFPN_{\uppercase\expandafter{\romannumeral2}}} (P_{3}^{\rm out}) ) )&\\
     &{\rm CCFPN_{\uppercase\expandafter{\romannumeral2}}} (P_{3}^{\rm out}) = {\rm Conv} ({\rm CCFPN_{\uppercase\expandafter{\romannumeral1}}} (P_{3}^{\rm out}) )&
\end{aligned}$}
\right.
\end{equation}

\begin{equation}
\label{EQ3}
\left\{\resizebox{0.85\hsize}{!}{$
\begin{aligned}
     &{\rm CCFPN_{\uppercase\expandafter{\romannumeral3}}} (P_{11}^{\rm in}) = {\rm Conv} ( P_{11}^{\rm in} ) &                      \\
     &{\rm CCFPN_{\uppercase\expandafter{\romannumeral3}}} (P_{12}^{\rm in}) = {\rm Conv} ( P_{12}^{\rm in}+ {\rm Up} ({\rm CCFPN_{\uppercase\expandafter{\romannumeral3}}} (P_{11}^{\rm in}) ) )&\\
     &{\rm CCFPN_{\uppercase\expandafter{\romannumeral3}}} (P_{13}^{\rm in}) = {\rm Conv} ( P_{12}^{\rm in}+ {\rm Up} (\rm CCFPN_{\uppercase\expandafter{\romannumeral3}}(P_{12}^{in}) ) )&\\
     &{\rm CCFPN_{\uppercase\expandafter{\romannumeral3}}} (P_{1}^{\rm out}) = {\rm Conv} ({\rm CCFPN_{\uppercase\expandafter{\romannumeral3}}} (P_{11}^{\rm in}) + P_{21}^{\rm in} ) &                      \\
     &{\rm CCFPN_{\uppercase\expandafter{\romannumeral3}}} (P_{2}^{\rm out}) = {\rm Conv} ({\rm CCFPN_{\uppercase\expandafter{\romannumeral3}}} (P_{12}^{\rm in}) + P_{22}^{\rm in} + {\rm Up} ({\rm CCFPN_{\uppercase\expandafter{\romannumeral3}}} (P_{1}^{\rm out}) ) )&\\
     &{\rm CCFPN_{\uppercase\expandafter{\romannumeral3}}} (P_{3}^{\rm out}) = {\rm Conv} ({\rm CCFPN_{\uppercase\expandafter{\romannumeral3}}} (P_{13}^{\rm in}) + P_{23}^{\rm in} + {\rm Up} ({\rm CCFPN_{\uppercase\expandafter{\romannumeral3}}} (P_{2}^{\rm out}) ) )&
\end{aligned}$}
\right.
\end{equation}

\begin{equation}
\label{EQ4}
\left\{\resizebox{0.85\hsize}{!}{$
\begin{aligned}
     &{\rm CCFPN_{\uppercase\expandafter{\romannumeral4}}} (P_{1}^{\rm out}) = {\rm Conv} ({\rm CCFPN_{\uppercase\expandafter{\romannumeral3}}} (P_{1}^{\rm out}) + {\rm Down} ({\rm CCFPN_{\uppercase\expandafter{\romannumeral4}}} (P_{2}^{\rm out}) )  ) &                      \\
     &{\rm CCFPN_{\uppercase\expandafter{\romannumeral4}}} (P_{2}^{\rm out}) = {\rm Conv} ({\rm CCFPN_{\uppercase\expandafter{\romannumeral3}}} (P_{2}^{\rm out})+ {\rm Down} ({\rm CCFPN_{\uppercase\expandafter{\romannumeral4}}} (P_{3}^{\rm out}) ) )&\\
     &{\rm CCFPN_{\uppercase\expandafter{\romannumeral4}}} (P_{3}^{\rm out}) = {\rm Conv} ({\rm CCFPN_{\uppercase\expandafter{\romannumeral3}}} (P_{3}^{\rm out}) )&
\end{aligned}$}
\right.
\end{equation}

In Eqs.~\ref{EQ1}-\ref{EQ4}, we unify the outputs of 
CCFPN-\uppercase\expandafter{\romannumeral1} $\sim$ CCFPN-\uppercase\expandafter{\romannumeral4} 
in Fig.~\ref{FIG:2} from top, middle to bottom into $( P_{1}^{\rm out}, P_{2}^{\rm out},P_{3}^{\rm out} )$, 
and add prefix ${\rm CCFPN_{\uppercase\expandafter{\romannumeral1}}}$ 
$\sim$ ${\rm CCFPN_{\uppercase\expandafter{\romannumeral4}}}$ to distinguish different CCFPN variants. 
For example, in Fig.~\ref{FIG:2}, ${\rm CCFPN_{\uppercase\expandafter{\romannumeral1}}} ( P_{1}^{\rm out})$ 
represents the top output in CCFPN-\uppercase\expandafter{\romannumeral1}, 
${\rm CCFPN_{\uppercase\expandafter{\romannumeral2}}}( P_{2}^{\rm out})$ represents the middle output in 
CCFPN-\uppercase\expandafter{\romannumeral2}, and 
${\rm CCFPN_{\uppercase\expandafter{\romannumeral3}}}( P_{3}^{\rm out})$ represents the bottom output in 
CCFPN-\uppercase\expandafter{\romannumeral3}. In Eqs.~\ref{EQ1}-\ref{EQ4}, the up and Down arrows represent 
an up-sampling and down-sampling operation for resolution matching, respectively, and Conv is a convolution 
operation for feature processing.

Unlike other FPNs~\cite{lin2017feature,liu2018path,tan2020efficientdet}, CCFPN uses more multi-scale input features 
$\mathbf{P}^{\rm  in}$ in a novel manner. From Fig.~\ref{FIG:2} and formulas (Eqs.~\ref{EQ1}-\ref{EQ4}), it can 
be found that CCFPN-\uppercase\expandafter{\romannumeral1} and CCFPN-\uppercase\expandafter{\romannumeral3} 
integrate more low-level detail information and more high-level semantic information, increasing the receptive 
field of the low-level, so that the low-level can obtain more context information when doing tiny object detection. 
CCFPN-\uppercase\expandafter{\romannumeral2} and CCFPN-\uppercase\expandafter{\romannumeral4} following 
CCFPN-\uppercase\expandafter{\romannumeral1} and CCFPN-\uppercase\expandafter{\romannumeral3} can shorten the 
information transmission path and use the accurate positioning information of low-level features. To sum up, 
in detecting sperms and impurities, CCFPN makes use of more contextual information and accurate positioning 
information about sperms and impurities to improve the detection rate.

\subsection{Head Network}
Three bounding boxes are predicted for each unit in the output feature map~\cite{redmon2018yolov3}. For each 
bounding box, seven coordinates ($t_{x}$, $t_{y}$, $t_{w}$, $t_{h}$, $C$, $P_{0}$, and $P_{1}$) are predicted, 
$t_{x}$ and $t_{y}$ are the offsets gained by the ACTIVE through sigmoid, and $t_{w}$ and $t_{h}$ are the 
scaling factors with the \textit{a priori} box, $C$ is the confidence level about whether an object exists in 
the bounding box, $P_{0}$ and $P_{1}$ represent the probability of sperms and impurities in the bounding box. 
For each cell, assume that the offset from the upper left corner of the image is ($C_{x}$, $C_{y}$), and the 
width and height of the corresponding \textit{a priori} box are $P_{w}$ and $P_{h}$. The calculation method of 
the center coordinates ($b_{x}$ and $b_{y}$), width ($b_{w}$) and height ($b_{h}$) of the prediction box is shown 
in Fig~\ref{FIG:2}. Multi-label classification is applied to predict the categories in each bounding box. 
In addition, because the head network adopts the dense prediction method, a non-maximum suppression method based 
on distance intersection on union set~\cite{zheng2020distance} is used to remove the boundary boxes with high 
overlap in the network output results.

\section{Experiment}\label{section:ea}
\subsection{Experimental Settings}
\subsubsection{DataSet}
In our work, our data are from subset-A of the SVIA dataset, a disclosed large-scale dataset for sperm 
detection~\cite{chen2022svia}. Subset-A contains 101 videos and 3622 sperm images (frames). We first group the 
sperm videos into the training, validation, and test sets with a ratio of 6:2:2. Then we use the images obtained 
by framing these videos. The exemplary images from SVIA Subset-A are shown in Fig.~\ref{FIG:3}.
\begin{figure}[!htbp]
	\centering
	\includegraphics[width=0.95\linewidth]{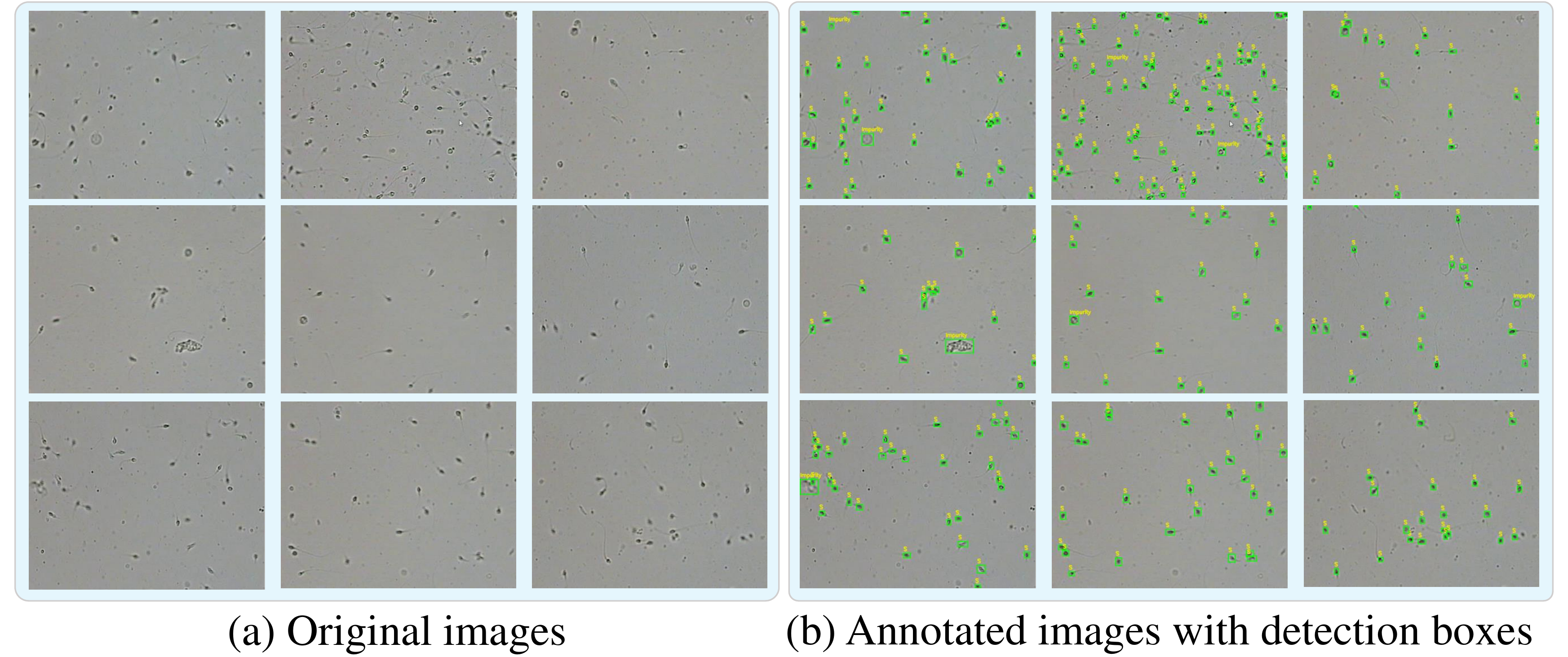}
	\caption{Examples of the images in SVIA Subset-A.}
	\label{FIG:3}
 \vspace{-0.2cm}
\end{figure}

\subsubsection{Implementation Details}
The experiment is conducted in Python 3.6.16 and Pytorch 1.7.1 in Windows 10. Our workstation is with 
Intel(R) Core(TM) i7-10700 CPU with 3.00GHz, 32GB RAM, and NVIDIA GEFORCE RTX 3060 12GB. For location and 
classification, we use the IOU function (location loss function) and the binary cross-entropy function 
(confidence and classification loss function)~\cite{redmon2018yolov3}. For optimization, we use Adam optimizer 
with default parameters for the whole experiment. For the other hyperparameters, the freeze training strategy 
is used in the experiments, when freezing partial layers and freezing no layers, the batch size is set to 4 and 2, 
the epoch number is 50 and 100, and the learning rate is set to $1\times 10^{-3}$ and $1\times 10^{-4}$, 
respectively. The input images are resized to 416 $\times$ 416 pixels.

\subsubsection{Evaluation Metrics}
${\rm AP}_{50}$ is used as a metric for performance evaluation. ${\rm AP}_{50}$ represents the value of AP 
when IOU is 0.5. The definitions of AP and IOU are: 

\begin{equation}
\label{EQ5}
\footnotesize
   \lvert\{
     \begin{aligned}  
     &{\rm Precision}=\frac{{\rm TP}}{{\rm TP} + {\rm FP}}, ~~~~~~~~~~~~~{\rm Recall}=\frac{{\rm TP}}{{\rm TP} + {\rm FN}}, \\ 
     &{\rm AP}=\frac{\sum_{i=1}^N \rm Precision_{\rm {max}}(\textit{i})}{\rm Number~of~Annotations},~{\rm IOU}=\frac{\rm G_{True} \bigcap G_{Pred}}{{\rm G_{True} \bigcup G_{Pred}}},&\\
    \end{aligned}
   \rvert
\end{equation}
where TP, TN, FP, and FN represent True Positive, True Negative, False Positive, and False Negative, $N$ denotes 
the number of detected objects. $G_{\rm True}$ and $G_{\rm Pred}$ represent the areas of the real box and the 
prediction box, respectively.

\subsection{Evaluation of Sperm and Impurity Detection Methods}
To demonstrate the effectiveness of the ACTIVE models for the detection of sperms and impurities in sperm 
microscopic images, we compare our model with the aforementioned deep learning detection methods including 
one-stage object detection (Faster-RCNN) and two-stage object detection (SSD, EfficientDet-D2 and YOLO-V3/V4). 
The experiments validate that our proposed model can outperform the existing advanced object detection models 
by a large margin.
\begin{figure}[!htbp]
	\centering
	\includegraphics[width=0.95\linewidth]{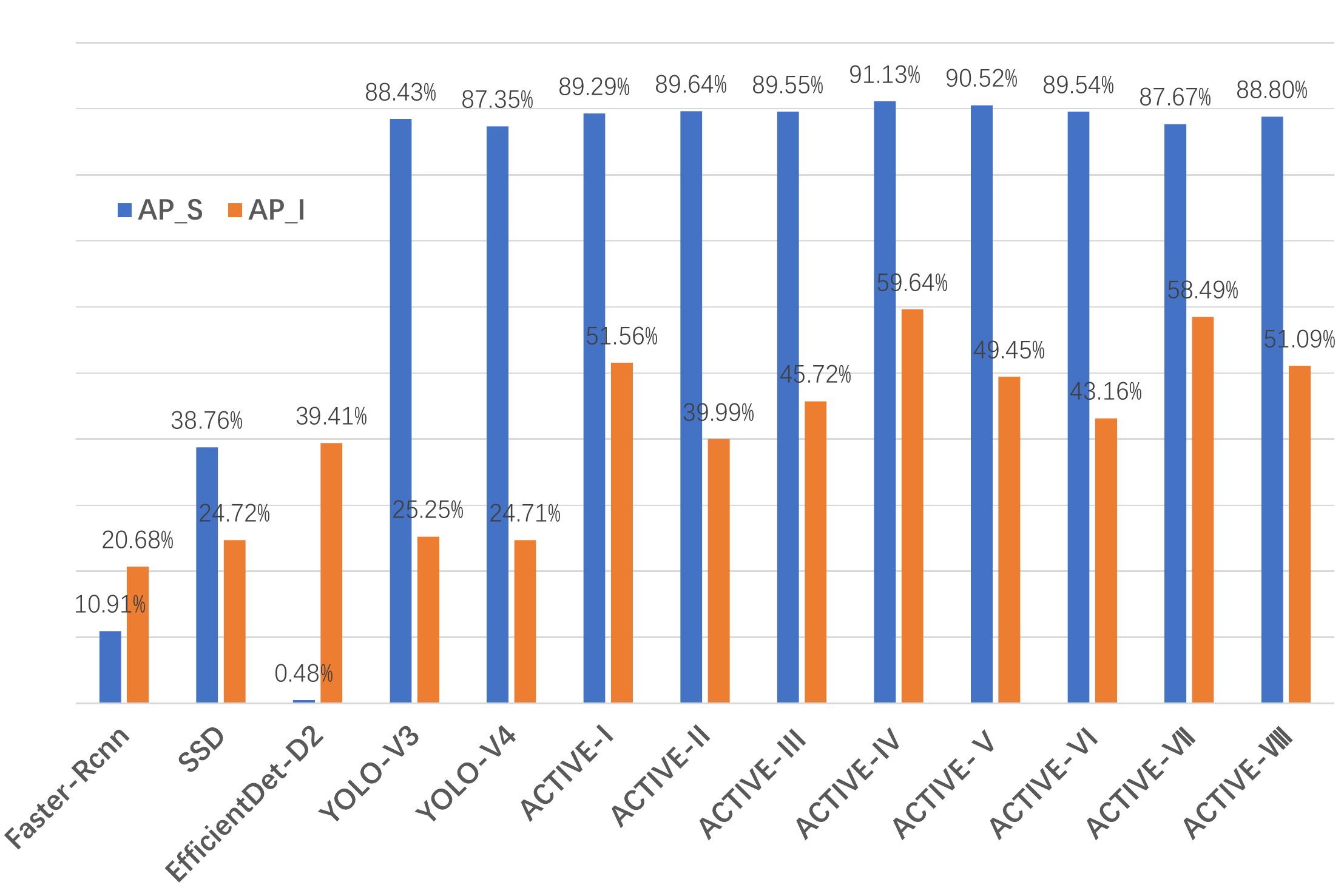}
	\caption{The AP of some deep learning detection methods and ACTIVE family models. 
 AP\_I and AP\_S represent AP of impurity and AP of sperm, respectively. 
 The highest performance of our proposed models obtains 91.13\% and 59.64\% of ${\rm AP}_{50}$ of the sperm and 
 impurity  detection with ACTIVE-IV.}
	\label{FIG:4}
 \vspace{-0.6cm}
\end{figure}

$\bullet$ \textbf{Evaluation of the sperm and impurity detection performance.} Fig.~\ref{FIG:4} shows the APs of 
all object detection methods. It can be seen that YOLO-V3 has good performance in sperm detection, but poor 
performance in impurity detection. EfficientDet-D2 performs satisfactorily in impurity detection, but its performance 
in sperm detection is poor. Compared with its competitors, the ACTIVE model family not only improves the performance 
of sperm detection but also greatly improves the performance of impurity detection. In particular, compared with 
YOLO-V3, which has high sperm detection performance, the AP of sperm detection of 
ACTIVE-\uppercase\expandafter{\romannumeral5} is increased by 2.7\%, while the AP of impurity detection of 
ACTIVE-\uppercase\expandafter{\romannumeral5} is increased by 34.39\%. Compared with EfficientDet-D2 with high 
impurity detection performance, the AP of sperm detection of ACTIVE-\uppercase\expandafter{\romannumeral5} is 
incredibly increased by 90.09\%, while the AP of impurity detection of ACTIVE-\uppercase\expandafter{\romannumeral5} 
is also greatly increased by 20.23\%. What's more, to visually appreciate the superior performance of ACTIVE, 
the detection results generated by the existing models and ACTIVE-\uppercase\expandafter{\romannumeral4} model 
are in Fig.~\ref{FIG:5}, wherein the ``Sperm-normal'' scenes (healthy) and the ``Sperm-lack'' scenes (oligospermia), 
the number of detection and correct detection cases of ACTIVE-\uppercase\expandafter{\romannumeral4} are the best, 
far higher than other deep models. In \textbf{Supplementary Materials}, we show that how our model can assist doctors in 
clinical sperm diagnosis better than SSD and YOLO-v4.

\begin{figure}[!htbp]
\includegraphics[width=0.95\linewidth]{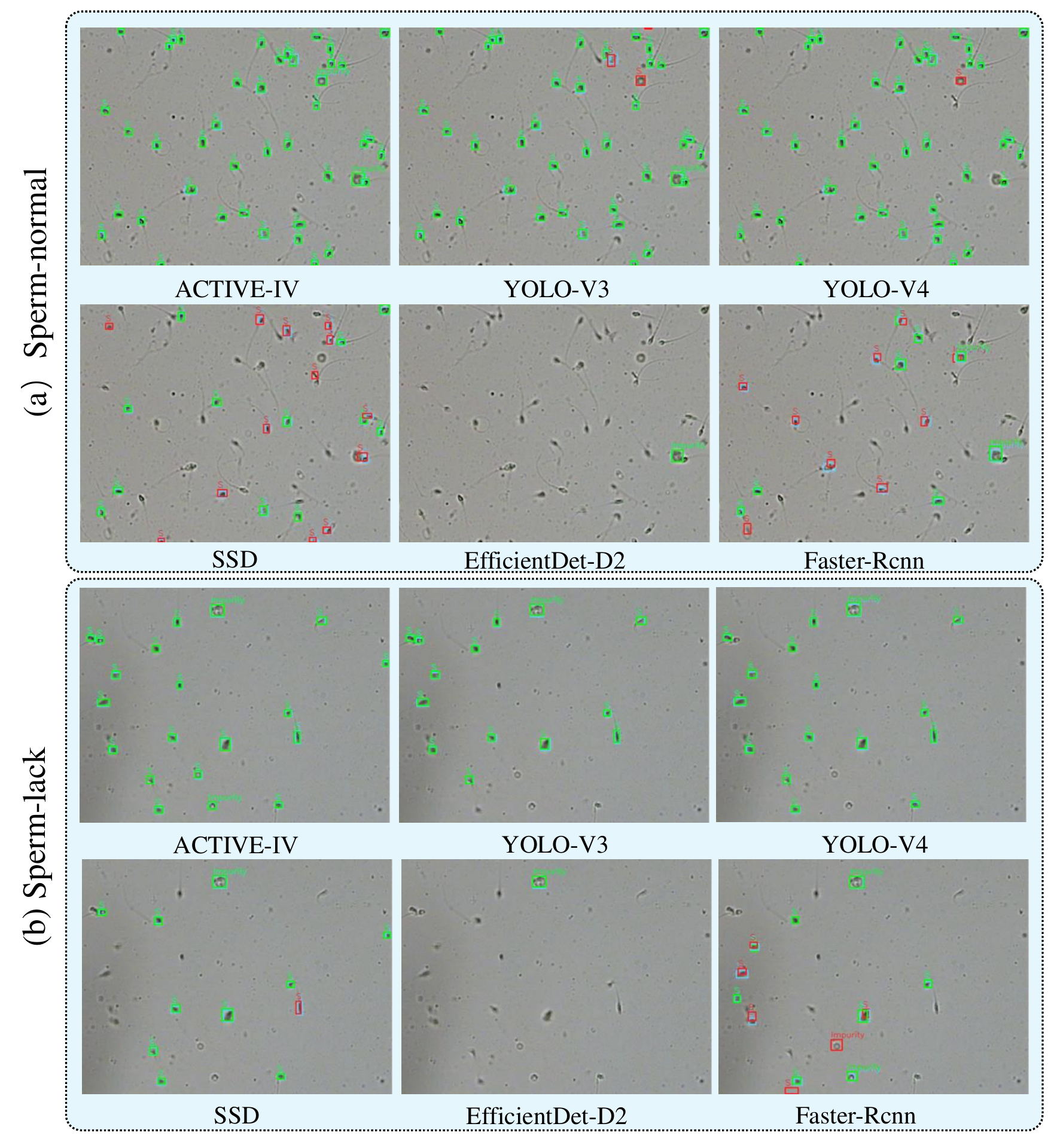}
\caption{Comparison of ACTIVE-\uppercase\expandafter{\romannumeral4} with YOLO-V4, YOLO-V3, SSD, EfficientDet-D2 and Faster-RCNN. 
The blue, green, and red boxes represent the GT, the correctly detected objects, and the incorrectly detected objects, respectively. }
\label{FIG:5}
\end{figure}

$\bullet$ \textbf{Cross-validation experiment.}
To evaluate the reliability and accuracy of ACTIVE, we conduct five-cross validation. The experimental results are 
shown in Tab.~\ref{tbl3}. It can be found that EfficinetDet-D2 obtains the best impurity detection performance and 
ACTIVE-\uppercase\expandafter{\romannumeral4} achieves the best sperm detection performance. Compared with 
EfficinetDet-D2, the AP value of impurities in ACTIVE series models is decreased by about 10 \%, but the AP value 
of sperm is increased by about 70\%. Compared with YOLO-V3/V4 with competitive sperm detection performance, the AP 
value of sperm in ACTIVE series models is increased by about 2\%, while the AP value of impurities is increased by 
about 10\%. In general, ACTIVE series models are superior to existing models in sperms and impurities detection. 
In Supplementary Materials, we also compare the memory, training cost, and frames per second (FPS) of different models.
\begin{table}[!htbp]
\centering
\captionsetup{justification=centering}
\caption{\\The detection results of the five-fold cross-validation experiments. 
( AP\_I and AP\_S respectively represent the average value of AP of sperms and impurities. )}
\label{tbl3}
\setlength{\tabcolsep}{8mm}{
\begin{tabular}{lll}
\hline
Model           & AP\_I (\%)     & AP\_S (\%)    \\
\hline
Faster-Rcnn     & 39.05 & 19.48 \\
SSD             & 3.59  & 40.95 \\
EfficientDet-D2 & \textbf{48.8}  & 22.04 \\
YOLO-V3         & 27.97 & 89.64 \\
YOLO-V4         & 28.75 & 87.92 \\
ACTIVE-\uppercase\expandafter{\romannumeral1}      & 41.03 & 90.24 \\
ACTIVE-\uppercase\expandafter{\romannumeral2}      & 35.18 & 90.72 \\
ACTIVE-\uppercase\expandafter{\romannumeral3}      & 39.55 & 90.93 \\
ACTIVE-\uppercase\expandafter{\romannumeral4}      & 37.18 & \textbf{91.23} \\
ACTIVE-\uppercase\expandafter{\romannumeral5}      & 36.34 & 91.03 \\
ACTIVE-\uppercase\expandafter{\romannumeral6}      & 32.39 & 90.61 \\
ACTIVE-\uppercase\expandafter{\romannumeral7}      & 36.39 & 90.43 \\
ACTIVE-\uppercase\expandafter{\romannumeral8}      & 33.24 & 90.18\\ \hline
\end{tabular}
}
\vspace{-0.3cm}
\end{table}

$\bullet$ \textbf{Ablation experiment.}
To evaluate the effectiveness of our proposed double branches in the ACTIVE framework, we conduct ablation 
experiments, as shown in Tab.~\ref{tbl2}. Compared with only using branch 1, the AP value of sperm and impurity 
using a double branch feature extraction network is increased by 0.85\% and 26.31\%, respectively, and the 
gain is substantial. Compared with only using branch 2, the AP value of sperm and impurities using a double 
branch feature extraction network is increased by 12.15\% and 88.81\%, respectively. Therefore, the DBFEN 
effectively improves the detection performance of sperms and impurities. This network is effective for the 
detection of sperms and impurities.
\begin{table}[htbp!]
\centering
\captionsetup{justification=centering}
\caption{\\Results of the ablation experiments. 
(DBFEN1 and DBFEN2 represent branch 1 and branch 2 in the DBFEN network respectively. 
\checkmark means it is used.)}
\label{tbl2}
\setlength{\tabcolsep}{6mm}{
\begin{tabular}{ccll}
\hline
\multicolumn{1}{l}{DBFEN1} & \multicolumn{1}{l}{DBFEN2} & Category & AP (\%)      \\ \hline
\multirow{2}{*}{\checkmark}         & \multirow{2}{*}{}          & Impurity   & 25.25 \\ \cline{3-4} 
                           &                            & Sperm    & 88.43 \\ \hline
\multirow{2}{*}{}          & \multirow{2}{*}{\checkmark}         & Impurity   & 39.41 \\ \cline{3-4} 
                           &                            & Sperm    & 0.48 \\ \hline
\multirow{2}{*}{\checkmark}         & \multirow{2}{*}{\checkmark}         & Impurity   & \textbf{51.56} \\ \cline{3-4} 
                           &                            & Sperm    & \textbf{89.29} \\ \hline
\end{tabular}
}
\vspace{-0.3cm}
\end{table}

\section{Discussion}
\label{section:dis}

\par
Here, we discuss the limitations of 
ACTIVE-(\uppercase\expandafter{\romannumeral1}-\uppercase\expandafter{\romannumeral8}) models for future improvement. 
As shown in Fig.~\ref{FIG:7}, there exist three kinds of imperfections in ACTIVE: overlooked, misclassified, and 
incomplete detection objects.
\begin{figure}[!htbp]
	\centering
	\includegraphics[width=0.98\linewidth]{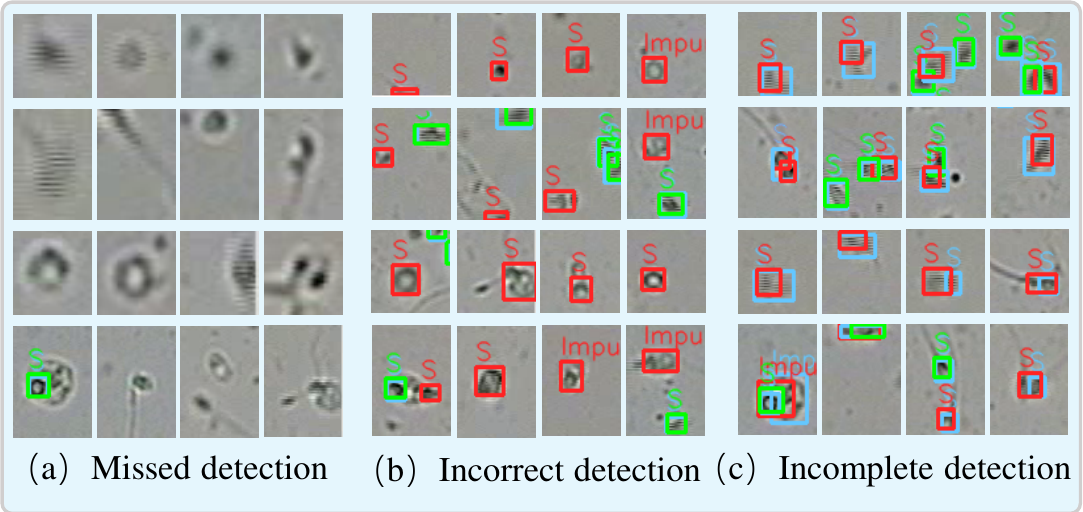}
	\flushleft
	\caption{Examples of error detection by the ACTIVE family models. 
 The blue, green, and red boxes represent the GT,  the correctly and incorrectly detected objects, respectively.}
 \vspace{-0.6cm}
	\label{FIG:7}
\end{figure}

\par
\textbf{i)} Fig.~\ref{FIG:7}(a) suggests that the reason why models overlook sperms and impurities is because 
of the following cases: low-resolution, residual shadow, and too tiny objects. 
The low-resolution object is due to the low quality of the data set, which may lose valuable information. 
The residual shadow object is generated because of the too fast-moving objects during the collection of sperm 
and impurity data, which is unavoidable but causes a streaked morphology. In addition, some objects are too 
tiny to be leveraged.

\textbf{ii)} The incorrect detection of objects, as shown in Fig.~\ref{FIG:7}(b), is mainly due to the identical 
sperm and impurity, overlapping of sperm and impurity positions, and the edge of the object position. It is 
challenging to explore the complete information for sperms and impurities that appear at the edges. Thus, these 
sperms and impurities are generally missed in the detection. Furthermore, to ensure the reliability of the 
annotated information, the dataset is only annotated with sperms and impurities that are not disputed. Several 
sperms and impurities may be deeply located in the wet film of semen. However, their microscopic images are not 
precise, and it is difficult to distinguish whether they are sperms or impurities, so they are not annotated in 
the dataset. Unannotated sperms and impurities can be detected, leading to false detection.

\textbf{iii)} As shown in Fig.~\ref{FIG:7}(c), the incomplete detection is because of uncertain morphology 
(double-headed sperm, large-headed sperm, etc.), low-resolution object, residual shadow object, overlapping of 
the sperm and impurity positions (a two-head sperm's features are identical to impurities or sperms overlapping), 
and the edge of the object position.

\vspace{-0.2cm}
\section{Conclusion and Future Work}
\label{section:c}
In this paper, we propose the model ACTIVE for the object detection task of microscopic sperm images using 
the DBFEN and the CCFPN. The proposed ACTIVE model aided by CCFPN has established the state-of-the-art sperm 
and impurity detection performance on the SVIA dataset. In the future, we plan to increase the number of 
annotations of microscopic sperm images in the SVIA dataset and optimize the memory cost of ACTIVE. Also, 
we will consider the use of the image super-resolution technique in microscopic sperm images to enhance 
sperm detection. 

\vspace{-0.3cm}


%

%

\ifCLASSOPTIONcaptionsoff
  \newpage
\fi

\bibliographystyle{IEEEtran}
\bibliography{Chenao}

\end{document}


%
\title{Supplementary Experiments of \\ 
``ACTIVE: A Deep Model for Sperm and Impurity Detection in Microscopic Video''}
%
%
%

\author{Ao~Chen,
        Jinghua~Zhang,
        Md Mamunur Rahaman,
        Hongzan~Sun, \textit{M.D.},
        Tieyong Zeng,
        Marcin Grzegorzek, 
        Feng-Lei~Fan$^*$, \textit{IEEE Member},
        Chen~Li$^*$
}

%
%

\markboth{Journal of \LaTeX\ Class Files,~Vol.~**, No.~**, May~2023}%
{Shell \MakeLowercase{\textit{et al.}}: Bare Demo of IEEEtran.cls for IEEE Journals\cite{iammarrone2003male}}
%



\maketitle


%
\IEEEpeerreviewmaketitle

\section{Results from Different Variants}

\begin{figure}[!htbp]
	\centering
	\includegraphics[width=0.98\linewidth]{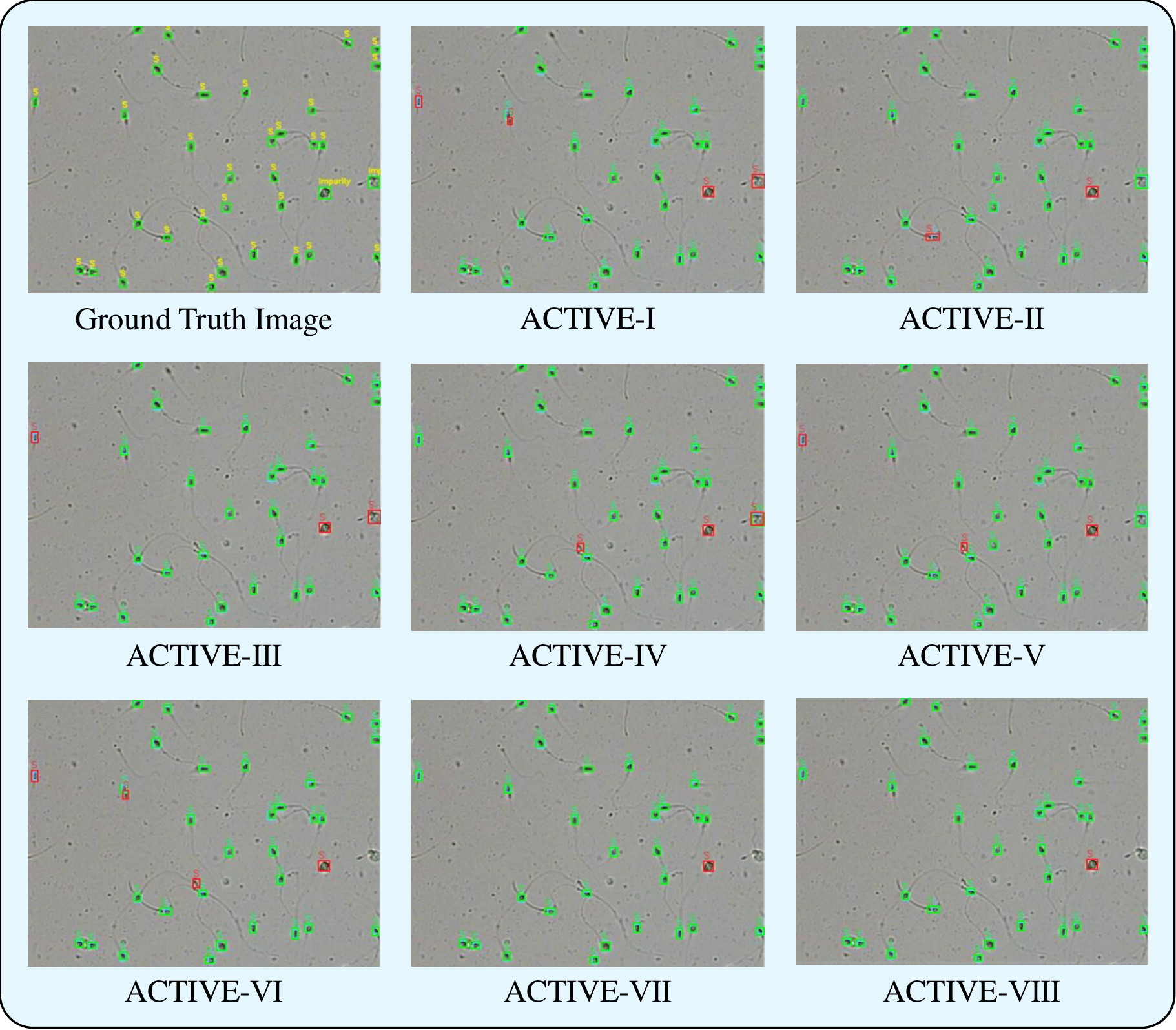}
	\flushleft
	\caption{GT images in comparison with the resultant images of ACTIVE models. In these images, 
 the blue, green, and red boxes represent the corresponding GT,  the correctly detected objects, 
 and the incorrectly detected objects, respectively.}
	\label{FIG:6}
\end{figure}

Fig.~\ref{FIG:6} summarizes ground-truth images in comparison with the resultant images of 
different ACTIVE model variants.

\section{Sperm Tracking and Motility Parameter Calculation}

\begin{figure}[!htbp]
\includegraphics[width=0.98\linewidth]{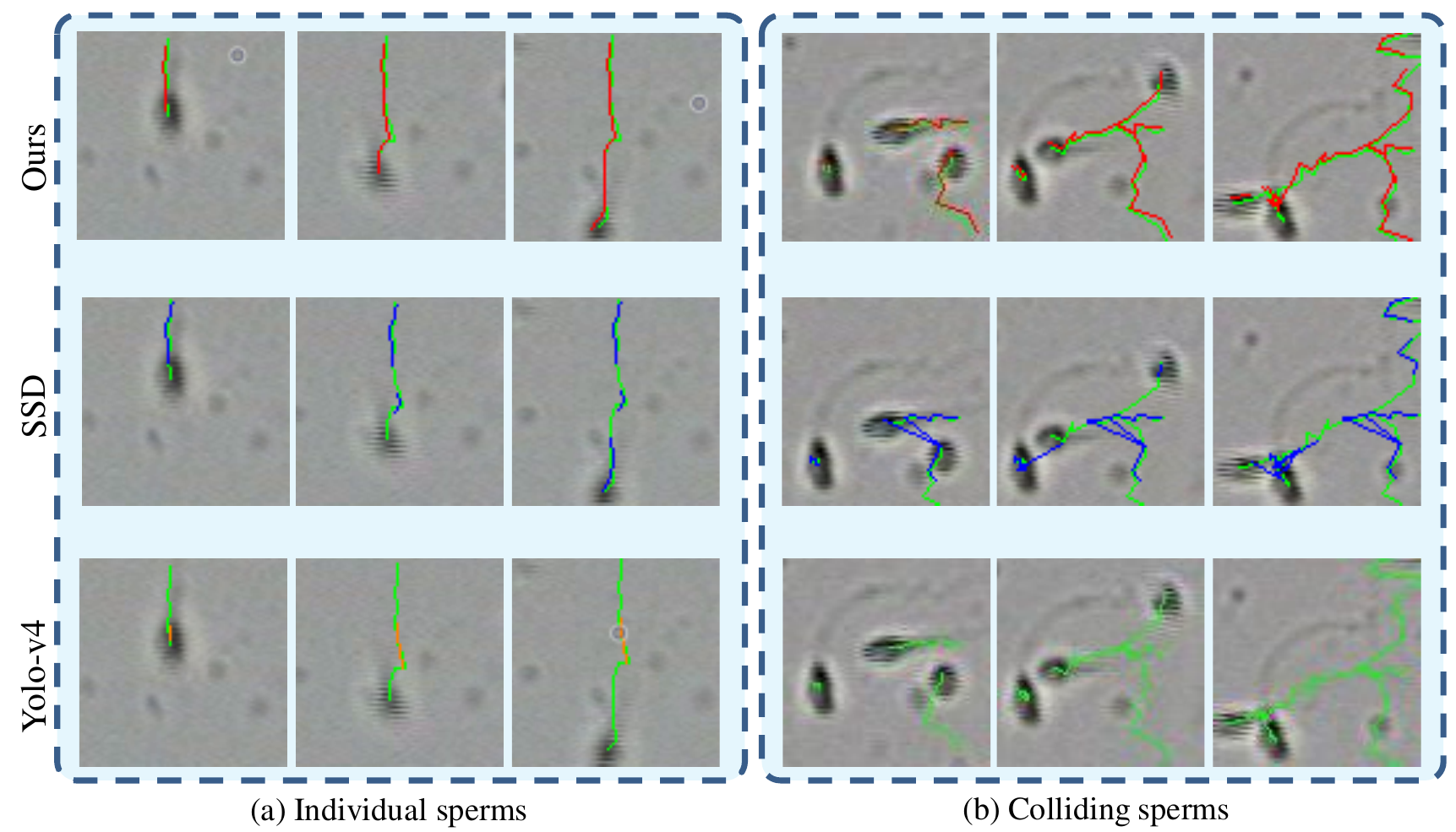}
\caption{A comparison of sperm tracking results.}
\label{FIG:spermtracking}
\end{figure}

The ultimate goal of sperm detection is to find sperm trajectories and calculate the relevant parameters for 
the diagnosis of sperm motility. The sperm tracking results of our method and two other models (YOLO-v4 and SSD) 
are compared in Fig.~\ref{FIG:spermtracking}. After the detection of sperms, kNN algorithm is applied to match 
sperms in the adjacent video frames~\cite{Li-2014-HSHDW} and find actual trajectories marked in SVIA Subset-B (more 
than 26,000 sperms in 10 videos for tiny object trackin tasks)~\cite{chen2022svia}. 
Then, three important motility parameters of sperms are calculated on Subset-B, including the Straight Line Velocity (VSL), 
Curvilinear Velocity (VCL), and Average Path Velocity (VAP)~\cite{Cooper-2010-WHORV,WHO-2021-WLMFT} according to the 
trajectories depicted using object detection algorithms. Compared with the actual trajectories, the error rates of VSL, 
VCL, and VAP calculated with ours (9.15\%, 4.59\%, and 7.95\%) are significantly lower than that of SSD (41.58\%, 
5.01\%, and 17.40\%) and Yolo-v4 (12.73\%, 36.12\%, and 19.65\%). Overall, our model can assist doctors in clinics 
better than SSD and YOLO-v4. Sperm morphology assessment has been one of the most common tests to assess fertility. 
Previous sperm morphology assessments have relied heavily on the laboratory physician to perform them manually, 
resulting in an extensive laboratory and interlaboratory variation~\cite{Wang2010-VITMA}. Our automated identification 
technology can help reduce this discrepancy, enabling clinicians to compare the results of tests performed at different 
times and locations.

\section{Evaluation of Memory, Time Costs and Frames Per Second (FPS)}

Tab.~\ref{tbl1} summarizes the memory costs, training time, and FPS of all models. From Tab.~\ref{tbl1}, 
it can be found that the ACTIVE family and YOLO family have more memory cost and training time than Faster-RCNN, SSD, 
and EfficientDet-2. This is because the ACTIVE family has an additional feature extraction branch, and uses a more 
complex multi-scale fusion approach. The memory costs of the ACTIVE family and YOLO family are by-and-large on the 
same level. In the ACTIVE family, the memory change caused by CCFPN is much more significant than adding more MBCB 
blocks. In particular, CCFPN-\uppercase\expandafter{\romannumeral1} and CCFPN-\uppercase\expandafter{\romannumeral5}'s 
memory and training time are similar. Overall, from the viewpoint of the greatly-enhanced detection performance and 
the benefits in medical diagnosis, the increased memory cost of our proposed method is acceptable.

\begin{table}[!htbp]
\centering
\captionsetup{justification=centering}
\caption{\\The memory costs, training time, and fps of some state-of-the-art models and ACTIVE family models. 
(DBFEN$n-n$ represents that branch 1 uses $n$ Residual Blocks and branch 2 uses $n$ MBCB Blocks in DBFEN.)}
\label{tbl1}
\scalebox{0.8}{
\begin{tabular}{@{}|l|l|l|l|l|l|@{}}
\hline
\multirow{2}{*}{Model} & \multirow{2}{*}{Backbone} & \multirow{2}{*}{Neck} & \multirow{2}{*}{\begin{tabular}[c]{@{}c@{}}Memory \\ Cost\end{tabular}} & \multirow{2}{*}{\begin{tabular}[c]{@{}c@{}}Training\\ Time\end{tabular}}
& \multirow{2}{*}{FPS} \\
                       &                           &                       &                              &                                &                        \\ \hline
                       \multirow{2}{*}{Faster-Rcnn} & \multirow{2}{*}{VGG16} & \multirow{2}{*}{RPN~\cite{girshick2015fast}} & \multirow{2}{*}{108MB} & \multirow{2}{*}{600min}
& \multirow{2}{*}{3.1} \\
                       &                           &                       &                              &                                &                        \\  \hline               
\multirow{2}{*}{SSD} & \multirow{2}{*}{VGG16} & \multirow{2}{*}{PFN~\cite{liu2016ssd}} & \multirow{2}{*}{91.1MB} & \multirow{2}{*}{143min}
& \multirow{2}{*}{35.0} \\
                       &                           &                       &                              &                                &                        \\  \hline
                       \multirow{2}{*}{EfficientDet-D2} & \multirow{2}{*}{EfficientNet-B2~\cite{tan2020efficientdet}} & \multirow{2}{*}{BiFPN~\cite{tan2020efficientdet}} & \multirow{2}{*}{31.2MB} & \multirow{2}{*}{528min}
& \multirow{2}{*}{22.8} \\
                       &                           &                       &                              &                                &                        \\  \hline
                       \multirow{2}{*}{YOLO-V4} & \multirow{2}{*}{CSPDarkNet53~\cite{bochkovskiy2020yolov4}} & \multirow{2}{*}{PANet} & \multirow{2}{*}{244MB} & \multirow{2}{*}{304min}
& \multirow{2}{*}{32.2} \\
                       &                           &                       &                              &                                &                        \\  \hline    
\multirow{2}{*}{YOLO-V3} & \multirow{2}{*}{Darknet-53} & \multirow{2}{*}{FPN} & \multirow{2}{*}{235MB} & \multirow{2}{*}{551min}
& \multirow{2}{*}{39.1} \\
                       &                           &                       &                              &                                &                        \\  \hline               
\multirow{2}{*}{ACTIVE-\uppercase\expandafter{\romannumeral1}} & \multirow{2}{*}{DBFEN23-23} & \multirow{2}{*}{CCFPN-\uppercase\expandafter{\romannumeral1}} & \multirow{2}{*}{327MB} & \multirow{2}{*}{567min}
& \multirow{2}{*}{33.4} \\
                       &                           &                       &                              &                                &                        \\  \hline
\multirow{2}{*}{ACTIVE-\uppercase\expandafter{\romannumeral2}} & \multirow{2}{*}{DBFEN23-23} & \multirow{2}{*}{CCFPN-\uppercase\expandafter{\romannumeral2}} & \multirow{2}{*}{475MB} & \multirow{2}{*}{581min}
& \multirow{2}{*}{28.6} \\
                       &                           &                       &                              &                                &                        \\  \hline
\multirow{2}{*}{ACTIVE-\uppercase\expandafter{\romannumeral3}} & \multirow{2}{*}{DBFEN23-23} & \multirow{2}{*}{CCFPN-\uppercase\expandafter{\romannumeral3}} & \multirow{2}{*}{473MB} & \multirow{2}{*}{647min}
& \multirow{2}{*}{30.4} \\
                       &                           &                       &                              &                                &                        \\  \hline                                          
                       \multirow{2}{*}{ACTIVE-\uppercase\expandafter{\romannumeral4}} & \multirow{2}{*}{DBFEN23-23} & \multirow{2}{*}{CCFPN-\uppercase\expandafter{\romannumeral4}} & \multirow{2}{*}{730MB} & \multirow{2}{*}{749min}
& \multirow{2}{*}{25.7} \\
                       &                           &                       &                              &                                &                        \\  \hline                     
\multirow{2}{*}{ACTIVE-\uppercase\expandafter{\romannumeral5}} & \multirow{2}{*}{DBFEN23-32} & \multirow{2}{*}{CCFPN-\uppercase\expandafter{\romannumeral1}} & \multirow{2}{*}{371MB} & \multirow{2}{*}{591min}
& \multirow{2}{*}{26.4} \\
                       &                           &                       &                              &                                &                        \\  \hline                     
\multirow{2}{*}{ACTIVE-\uppercase\expandafter{\romannumeral6}} & \multirow{2}{*}{DBFEN23-32} & \multirow{2}{*}{CCFPN-\uppercase\expandafter{\romannumeral2}} & \multirow{2}{*}{540MB} & \multirow{2}{*}{650min}
& \multirow{2}{*}{23.9} \\
                       &                           &                       &                              &                                &                        \\  \hline                     
\multirow{2}{*}{ACTIVE-\uppercase\expandafter{\romannumeral7}} & \multirow{2}{*}{DBFEN23-32} & \multirow{2}{*}{CCFPN-\uppercase\expandafter{\romannumeral3}} & \multirow{2}{*}{517MB} & \multirow{2}{*}{645min}
& \multirow{2}{*}{25.1} \\
                       &                           &                       &                              &                                &                        \\  \hline                     
\multirow{2}{*}{ACTIVE-\uppercase\expandafter{\romannumeral8}} & \multirow{2}{*}{DBFEN23-32} & \multirow{2}{*}{CCFPN-\uppercase\expandafter{\romannumeral4}} & \multirow{2}{*}{742MB} & \multirow{2}{*}{753min}
& \multirow{2}{*}{21.8} \\
                       &                           &                       &                              &                                &                        \\  \hline                     
\end{tabular}
}
\end{table}

\section{Declaration of Competing Interest}
The authors declare that they have no known competing financial interests or personal relationships that could 
have appeared to influence the work reported in this paper.

\section{Data and Ethical Statement }
All data used in this paper are from an open-source database SVIA (available at: https://github.com/Demozsj/Detection-Sperm), 
and there are no human or animal experiments in this paper.

\ifCLASSOPTIONcaptionsoff
  \newpage
\fi

\bibliographystyle{IEEEtran}
\bibliography{Chenao}
